\documentclass{article}

\usepackage{PRIMEarxiv}

\usepackage[utf8]{inputenc} 
\usepackage[T1]{fontenc}    
\usepackage{hyperref}       
\usepackage{url}            
\usepackage{booktabs}       
\usepackage{amsfonts}       
\usepackage{nicefrac}       
\usepackage{microtype}      
\usepackage{lipsum}
\usepackage{amsmath}
\usepackage{graphicx}
\graphicspath{{media/}}     
\usepackage{graphicx}
\usepackage{algorithm}      
\usepackage{algpseudocode} 
\usepackage{mdframed}
\DeclareMathOperator*{\argmaxA}{arg\,max} 
\usepackage{mdframed}
\newcommand{\indep}{\perp \!\!\! \perp}

\title{Box It to Bind It: Unified Layout Control and Attribute Binding in T2I Diffusion Models
}

\author{
  Ashkan Taghipour\\
  University of Western Australia \\
  \texttt{ashkan.taghipour@research.uwa} \\
   \And
  Morteza Ghahremani \\
  Munich Center for Machine Learning \\
  (MCML)\\
  \AND
  Mohammed Bennamoun \\
  University of Western Australia \\
  \And
  Aref Miri Rekavandi \\
  The University of Melbourne \\
  \And
  Hamid Laga \\
  Murdoch University \\
  \And
  Farid Boussaid \\
  University of Western Australia \\
}

\begin{document}
\maketitle

\begin{abstract}
While latent diffusion models (LDMs) excel at creating imaginative images, they often lack precision in semantic fidelity and spatial control over where objects are generated. To address these deficiencies, we introduce the Box-it-to-Bind-it (B2B) module - a novel, training-free approach for improving spatial control and semantic accuracy in text-to-image (T2I) diffusion models. 
B2B targets three key challenges in T2I: catastrophic neglect, attribute binding, and layout guidance. 
The process encompasses two main steps: i) \emph{Object generation}, which adjusts the latent encoding to guarantee object generation and directs it within specified bounding boxes, and ii) \emph{attribute binding}, guaranteeing that generated objects adhere to their specified attributes in the prompt.
B2B is designed as a compatible plug-and-play module for existing T2I models, markedly enhancing model performance in addressing the key challenges. 
We evaluate our technique using the established CompBench and TIFA score benchmarks, demonstrating significant performance improvements compared to existing methods. The source code will be made publicly available at \url{https://github.com/nextaistudio/BoxIt2BindIt}.
\end{abstract}


\begin{figure*}
    \centering
    \includegraphics[width=0.99\textwidth]{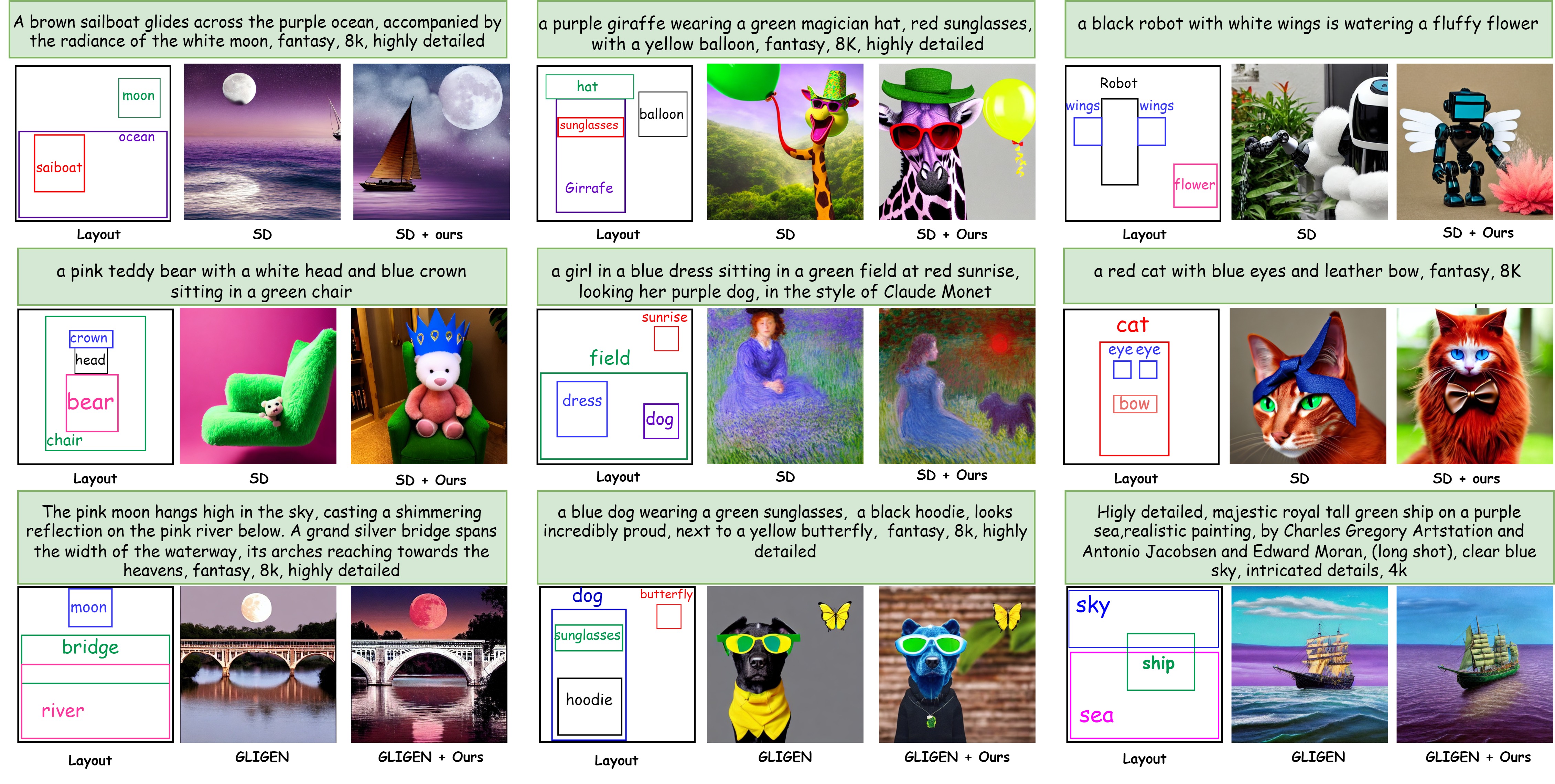}
    \caption{The proposed Box-it-to-Bind-it (B2B) is a training-free, plug-and-play tool. It is designed to enhance the performance of latent diffusion models (LDMs) such as Stable Diffusion~\cite{rombach2022highresolution} and GLIGEN~\cite{li2023gligen}. Its primary function is to improve the generation of objects and then bind their attributes within a specified layout.
    }
    \label{fig:image_cover}
\end{figure*}%

\section{Introduction}\label{sec:intro}
Diffusion models have shown impressive ability in generating both imaginative and realistic images \cite{mardani2023variational,sun2023dreamsync,peng2023metadataconditioned,Feng_2023_ICCV}. However, as shown in Figure \ref{fig:image_cover}, their ability to faithfully adhere to given prompts, especially in terms of object attributes \cite{huang2024creativesynth, yang2024mastering, yu2022scaling}, and to control object placement within specified bounding boxes, is limited \cite{chen2023trainingfree, kim2023dense}. This lack of spatial precise control and attribute binding in image generation highlights a key challenge in existing models \cite{lee2024compose,wang2024instantid}.
Thus far, two main categories of methods have been devised to tackle spatial control and semantic binding in latent diffusion models (LDMs): (a) the first involves either training a model from scratch or fine-tuning an existing diffusion model for a specific purpose, such as conditioning generation on additional inputs like pose, masks, etc \cite{li2023gligen,lee2024parrot, phung2023grounded, zhang2023adding}. 
Such methods typically require significant computational resources and a prolonged development period due to the increasing size of models and training datasets; (b) utilizing a pre-trained model and then integrating features that facilitate controlled generation without the need for extensive training or fine-tuning \cite{feng2023trainingfree,mo2023freecontrol,zhao2023loco, yu2023freedom}. 

The objective of this study is to propose a training-free approach called Box-it-to-Bind-it (B2B), addressing three key challenges T2I generation: 1) \textbf{Catastrophic Neglect}, which arises when one or more tokens, also referred to as objects, in a prompt are not generated \cite{chefer2023attendandexcite} 
\textbf{Attribute Binding} issues, occurring when the model either fails to correctly associate object attributes or mistakenly binds those to the wrong token (object)  
\cite{huang2023t2icompbench} 
, and 3) \textbf{Layout Guidance},  which focuses on guiding the diffusion process to generate objects within specified bounding boxes, enhancing spatial control.
The proposed B2B is designed to guide the LDMs' latent encoding in two steps during the inference phase: \emph{object generation} and \emph{attribute binding}. 
The former leverages Large Language Models (LLMs) to determine a reasonable layout for the given prompt, directing the diffusion process to precisely generate objects within specified bounding boxes. The latter is designed to ensure the effective binding of attributes to their respective objects. 
Our proposed framework is built as a plug-and-play module that is capable of indentation with any T2I generative model. 
In summary, the key contributions of this study include:
\begin{itemize}
    \item We propose a two-stage, training-free approach that significantly enhances the content of existing text-to-image (T2I) generative models. Our method adds precise spatial control and ensures faithful adherence to object attributes, effectively addressing catastrophic neglect and attribute-binding challenges.
    \item We formalize the \emph{object generation} and \emph{attribute binding} within a probabilistic model and design them as plug-and-play modules to bolster layout control and attribute binding in T2I models like Stable Diffusion~\cite{rombach2022highresolution} and layout-trained Gligen~\cite{li2023gligen}. 
    \item To show our method's superiority in attribute binding and spatial reasoning, we examined it through several comparative analyses and assessed its results by widely-used TIFA~\cite{hu2023tifa} and CompBench~\cite{huang2023t2icompbench} benchmarks and scores. 
\end{itemize}

\section{Related Works}\label{sec::related}
\begin{figure*}[t] 
\centering
\includegraphics[width=0.99\textwidth]{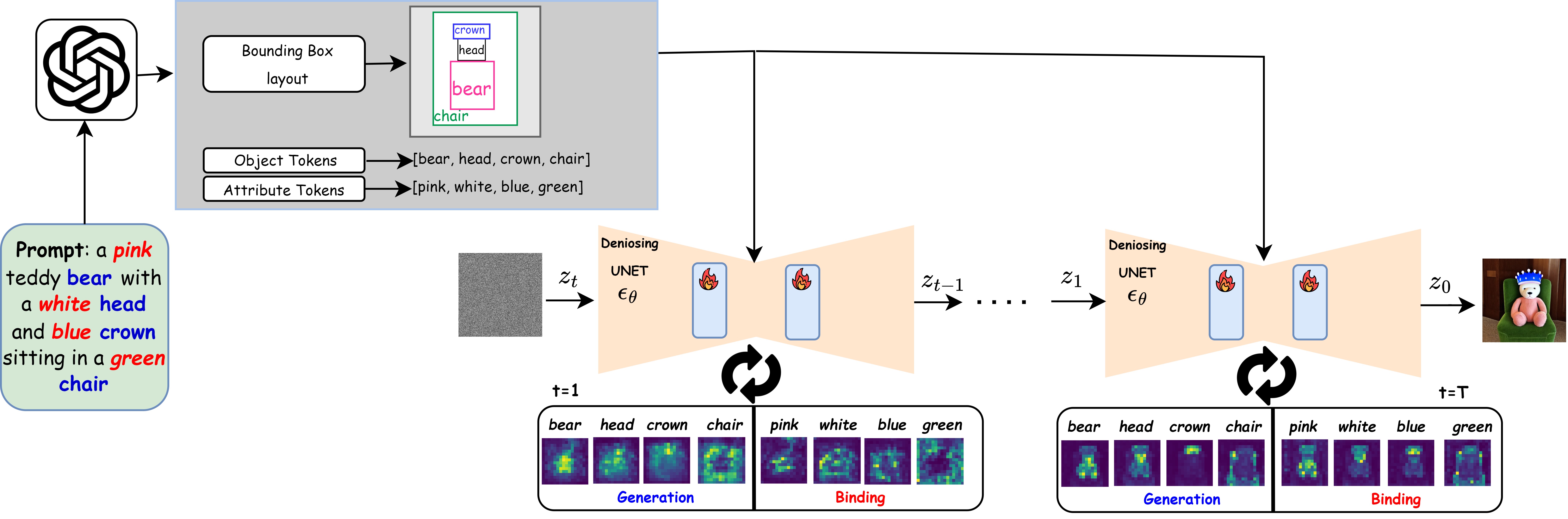} 
\caption{The framework of the proposed B2B method. Given a prompt, it first enters an LLM (here GPT-4) to extract the corresponding bounding box coordinates for each object in the text, the object tokens, and their respective attributes. In the latent space, this information is fed into the $16 \times 16$ cross-attention layer of the denoising UNet at specified timesteps $\mathcal{T}_t$. The \emph{generation} module ensures the generation of each object in the prompt and adherence to each object in the given layout while the \emph{binding} module is applied for attribute binding.}
\label{fig:image-framework}
\end{figure*}

\textbf{Image generation with diffusion models} 
Diffusion models are renowned for their ability to be trained using a variety of directive inputs, including text, layout, and category labels \cite{gupta2023photorealistic, basu2023editval, zhang2023adding}.
This versatility facilitates the effective generation of images conditioned on such inputs. However, challenges persist in generating images that accurately reflect the semantic meaning of a given prompt~\cite{yang2024mastering, Zeng_2023_CVPR, gani2023llm}. Additionally, current T2I diffusion models often struggle with spatial reasoning~\cite{wu2023selfcorrecting, zhang2023controllable, lian2023llmgrounded}, which limits their precision in positioning objects at specific locations.

In response, numerous methods have been developed in T2I diffusion models. Composable Diffusion~\cite{liu2022compositional} stands out as a pioneer training-free model, proposing the integration of multiple diffusion models during inference to ensure generated images encompass all concepts described in a prompt. However, it lacks semantic binding and control over the spatial location of generated objects. Divide\&Bind \cite{li2023divide} reduces catastrophic neglect and improves color binding with a total variation-based loss but struggles with rare colors and lacks layout control. Syngen \cite{rassin2023linguistic} employs linguistic identifiers to enhance the correlation between object attributes and corresponding tokens, but it lacks control over the location of generated objects, and the process of identifying linguistic identifiers for each prompt can be time-consuming, especially for lengthy prompts. 
The Attend-and-Excite method \cite{chefer2023attendandexcite} effectively addresses catastrophic neglect and attribute binding through an optimization function at the inference stage, yet it fails in spatial reasoning. 
Structured Diffusion Guidance \cite{feng2023trainingfree} incorporates language structures into diffusion guidance to mitigate catastrophic neglect but falls short in layout control. 
GORS \cite{huang2023t2icompbench} proposes fine-tuning the SD model \cite{rombach2022highresolution} using images closely aligned with compositional prompts, with the fine-tuning loss weighted by a reward alignment score. While this model is capable of generating plausible images, it lacks control over the locations of generated objects. 
A forward-backward guidance in the diffusion process is proposed in \cite{chen2023trainingfree} to ensure adherence of the generated image to both the given text and layout. However, it is primarily developed for zero-shot editing applications and does not focus on ensuring attribute binding. MultiDiffusion \cite{bartal2023multidiffusion} introduces a framework that fuses multiple diffusion processes for a controllable generation. Although this approach enables spatial reasoning, it is primarily designed for panorama image generation and manipulating the aspect ratio of generated images \cite{liu2023syncdreamer}. It is not applicable for prompts that contain several objects and attributes. Following this, BoxDiff \cite{Xie_2023_ICCV} introduced a box-constrained loss at the inference stage to improve layout control in the SD model. However, it faces challenges in effectively managing attributes when added to object tokens. \cite{gani2023llm} is a concurrent work that leverages LLMs for generating layouts and guiding the diffusion model to follow these layouts. The primary focus of this work, however, is on ensuring the generation of all objects in long prompts, rather than attribute binding. To this end, our method aims to bridge the gap between layout-based and attribute-based methods, ensuring that both the given layout and object attribute binding are satisfactorily followed

\section{Proposed Method}
\label{sec::method}
B2B is a reward-guided diffusion model that guides diffusion models toward the given text during inference. 
LDMs' sampling employs a decoder to generate an RGB image $\mathbf{x}\in \mathbb{R}^{H\times W\times3}$ from the latent space $\mathbf{z}\in \mathbb{R}^{C\times h\times w}$ that is conditioned on an input guiding text $\mathbf{y}$. 
Rombach et al.~\cite{rombach2022high} incorporate the cross-attention mechanism into their foundational UNet backbone to enhance the flexibility of conditional image generators. 
Let $\mathbf{\epsilon}_{\theta} (\mathbf{x}_t,t,y)$, $t\in\{1,\cdots,T\}$, represent a sequence of denoising UNets with gradients $\nabla {\theta}$  over a batch. 
The conditional LDM is learned through $T$ steps via
\begin{align}\label{eq:LDM}
L_{LDM}=\mathop{\mathbb{E}}_{\mathbf{z},\mathbf{y},\mathbf{\epsilon}\sim\mathcal{N}(\mathbf{0}, \mathbf{I})}\left[|| \mathbf{\epsilon} - \mathbf{\epsilon}_{\theta}(\mathbf{z}_t,t,\tau_{\theta}(y))||_2^2
    \right],
\end{align}
where $\tau_{\theta}(y)$ is a learnable encoder that projects text prompt $\mathbf{y}$ to an intermediate representation. Additional information about the training process can be found in \cite{rombach2022high,ho2020denoising}. 
For a text prompt with $L$ tokens, the attention-based transformer model yields an attention map of size $\mathbf{A}\in \mathbb{R}^{L\times h_{a}\times w_{a}}$, where $h_a$ and $w_a$ denote the height and width of the attention map, respectively.
In the given cross-attention map, $n_o$ objects with $n_a$ attributes $\mathbf{A}_S=\{\mathbf{A}_{o}^i\}_{i=1}^{n_o} \cup \{\mathbf{A}_{a}^j\}_{j=1}^{n_a}$ are a subset of $\mathbf{A}$ ($n_{o}+n_{a}\leq L$). 
The attributes can be color, texture, etc. The assignment of attributes is adaptable, allowing for the allocation of any arbitrary number of distinctive features to an object.

The proposed B2B operates within a zero-shot learning setting during inference that (\textit{i}) generates an image with $n_o$ objects specified in the text, and (\textit{ii}) ensures alignment between the objects and their corresponding attributions if they exist.
Given a fixed latent, we guide the latent encoding using the probabilistic model and the Bayesian approach as follows: 
\begin{align}\label{eq:Bayesian}
    \nonumber p\left(\mathbf{z}_t,\mathbf{A}_{o}^1,\cdots,\mathbf{A}_{o}^{n_o},\mathbf{A}_{a}^1,\cdots,\mathbf{A}_{a}^{n_a} | \nabla{\theta} \right)  &  \\\nonumber
    \propto p\left(\mathbf{z}_t,\mathbf{A}_{o}^1,\cdots,\mathbf{A}_{o}^{n_o},\mathbf{A}_{a}^1,\cdots,\mathbf{A}_{a}^{n_a}, \nabla{\theta} \right)\\
    =p\left(\nabla{\theta} | \mathbf{z}_t \right) p\left(\mathbf{A}_{o}^1,\cdots,\mathbf{A}_{o}^{n_o},\mathbf{A}_{a}^1,\cdots,\mathbf{A}_{a}^{n_a}| \mathbf{z}_t\right)
    &p\left(\mathbf{z}_t\right).
\end{align}
This equation introduces a Markov chain, wherein an enhanced $\mathbf{A}_S$ improves $\mathbf{z}_t$, subsequently improving $\nabla\theta$. 
Since we aim at guiding the attention maps of attributes toward their corresponding objects, equation~\ref{eq:Bayesian} can be simplified as 

\begin{align}\label{eq:obj_log}
    \nonumber
    \argmaxA_{\mathbf{z}_t,\mathbf{A}_S}\:\: &\text{log}\;p\left(\nabla{\theta} | \mathbf{z}_t \right) &\\  \nonumber
     &+\text{log}\;p\left(\mathbf{A}_{o}^1,\cdots,\mathbf{A}_{o}^{n_o},\mathbf{A}_{a}^1,\cdots,\mathbf{A}_{a}^{n_a}| \mathbf{z}_t\right)   &\\
     &+\text{log}\;p\left(\mathbf{z}_t\right)
\end{align}
Here, the first term implies the recovered latent encoding, accomplished by the LDM model. 
Likewise, the third term implies the prior latent encoding that is handled by the LDM model.  
The second term implies that the attention map should be consistent with the latent encoding as they interact to generate the same content. 
With this description, our method must optimize the simplified version of equation~\ref{eq:obj_log}:
\begin{equation}\label{eq:obj_final}
     \argmaxA_{\mathbf{A}_S}\:\: \text{log}\;p\left(\mathbf{A}_{o}^1,\cdots,\mathbf{A}_{o}^{n_o},\mathbf{A}_{a}^1,\cdots,\mathbf{A}_{a}^{n_a}| \mathbf{z}_t\right).
\end{equation}

Since objects in a scene are independent
($\mathbf{A}_{o}^i\mathbf{A}_{o}^l$), 
($\mathbf{A}_{o}^i\indep\mathbf{A}_{o}^l,\:\: i,l\in\{1,\cdots,n_o\},\:i\neq l$), the above equation can be rewritten in terms of pairs of objects and their attributes as follows:
\begin{equation}\label{eq:independent1}
     \argmaxA_{\mathbf{A}_S}\:\: \text{log}\;
     \prod_{\substack{i=<{n_o}>\\j=<{n_a}>}} p\left(\mathbf{A}_{o}^i,\mathbf{A}_{a}^j| \mathbf{z}_t\right).
\end{equation}
Applying the Bayes' rule to equation~\ref{eq:independent1} yields
\begin{align}\label{eq:independent2}
     \argmaxA_{\mathbf{A}_S}\:\: \text{log}
     &\prod_{\substack{i=<{n_o}>\\j=<{n_a}>}} \frac{p\left( \mathbf{z}_t|\mathbf{A}_{o}^i,\mathbf{A}_{a}^j\right)
     p\left(\mathbf{A}_{o}^i,\mathbf{A}_{a}^j\right)}{p(\mathbf{z}_t)}.
\end{align}
From equations~\ref{eq:independent1} and~\ref{eq:independent2}, it can be concluded that $p\left(\mathbf{A}_{o}^i,\mathbf{A}_{a}^j| \mathbf{z}_t\right) \propto  p\left(\mathbf{A}_{o}^i,\mathbf{A}_{a}^j\right)$, hence

\begin{figure}[t] 
\centering
\includegraphics[width=0.6\columnwidth]{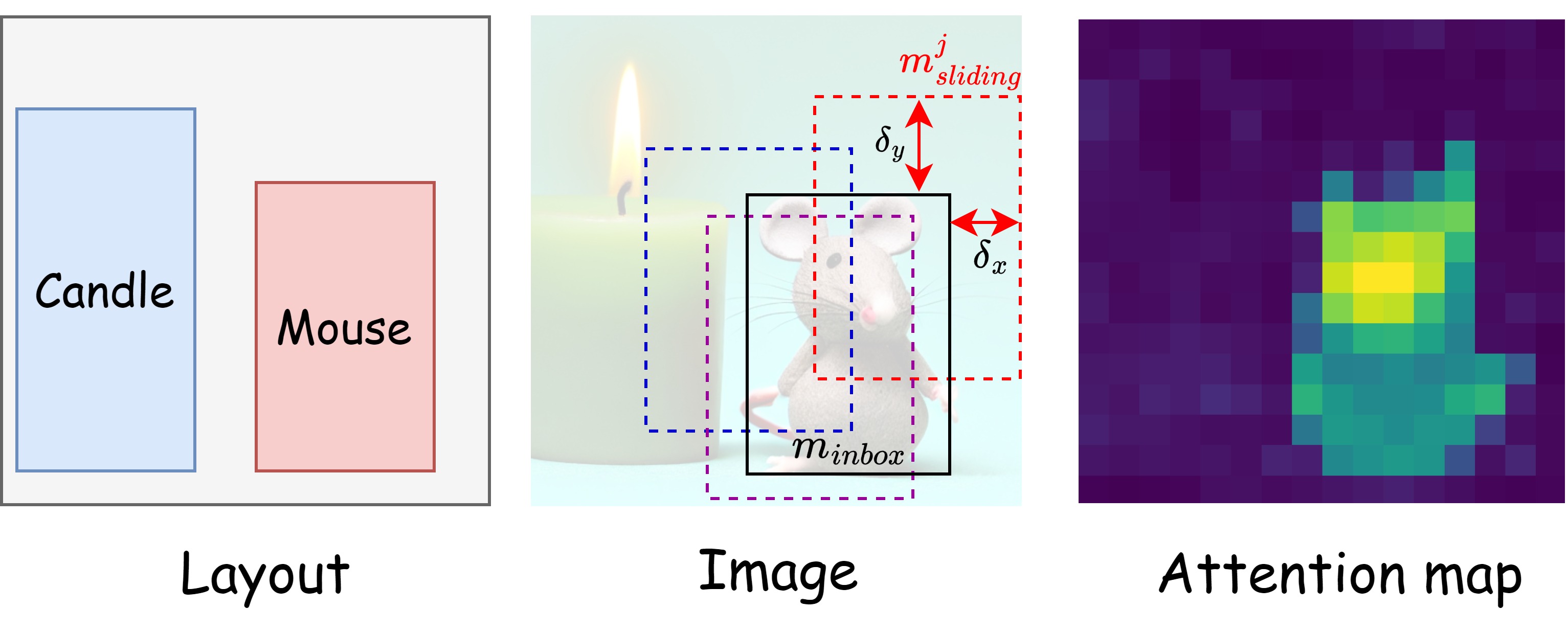}
\caption{IoU-based framework for object generation. 
As LDMs do not generally position objects within their designated bounding boxes, we enforce LDMs to generate objects centered within the specified bounding box by exerting additional $N$ boxes that push them away from the borders.
}
\label{fig:IoU-framework}
\end{figure}

\begin{equation}\label{eq:final}
     \argmaxA_{\mathbf{A}_S}\:\: \text{log}
     \prod_{\substack{i=<n_o>\\j=<n_a>}} p\left(\mathbf{A}_{a}^j|\mathbf{A}_{o}^i\right)
     p\left(\mathbf{A}_{o}^i\right).
\end{equation}

The equation indicates that we \textit{i} must increase the probability of every object's presence (or object's generation) in a scene, \textit{ii} simultaneously, its associated attributes should also be maximized. Note that one could theoretically condition objects on attributes, this assumption is unreasonable in practice as our goal is to bind attribute features with their corresponding objects.

We increase the probability of generated objects and their corresponding attributes via a reward-based approach, introduced in studies~\cite{yuan2023rewarddirected, huang2023diffusion}. 
We decompose equation~\ref{eq:final} into \emph{object reward} $\mathcal{R}_o$ and \emph{attribute reward} $\mathcal{R}_a$, compute both in every step, and then update the latent $\mathbf{z}_{t}$
\begin{align}\label{eq:finalgrad}
     \mathbf{z}_{t}^{\prime} = \mathbf{z}_{t}+\gamma \nabla\left(\sum_{i=1}^{n_o}\mathcal{R}_o^{i}+\lambda_{a}\sum_{j=1}^{n_a}\mathcal{R}_a^{j}\right),
\end{align}

\noindent where $\gamma$ is a hyperparameter that determines the reward step.
Likewise, $\lambda_{a}$ is a hyperparameter that makes a balance between the object and attribute rewards. 
In the following, we detail the reward computation for the objects and attributes. 

\begin{figure}[t]
\centering
\includegraphics[width=0.55\columnwidth]{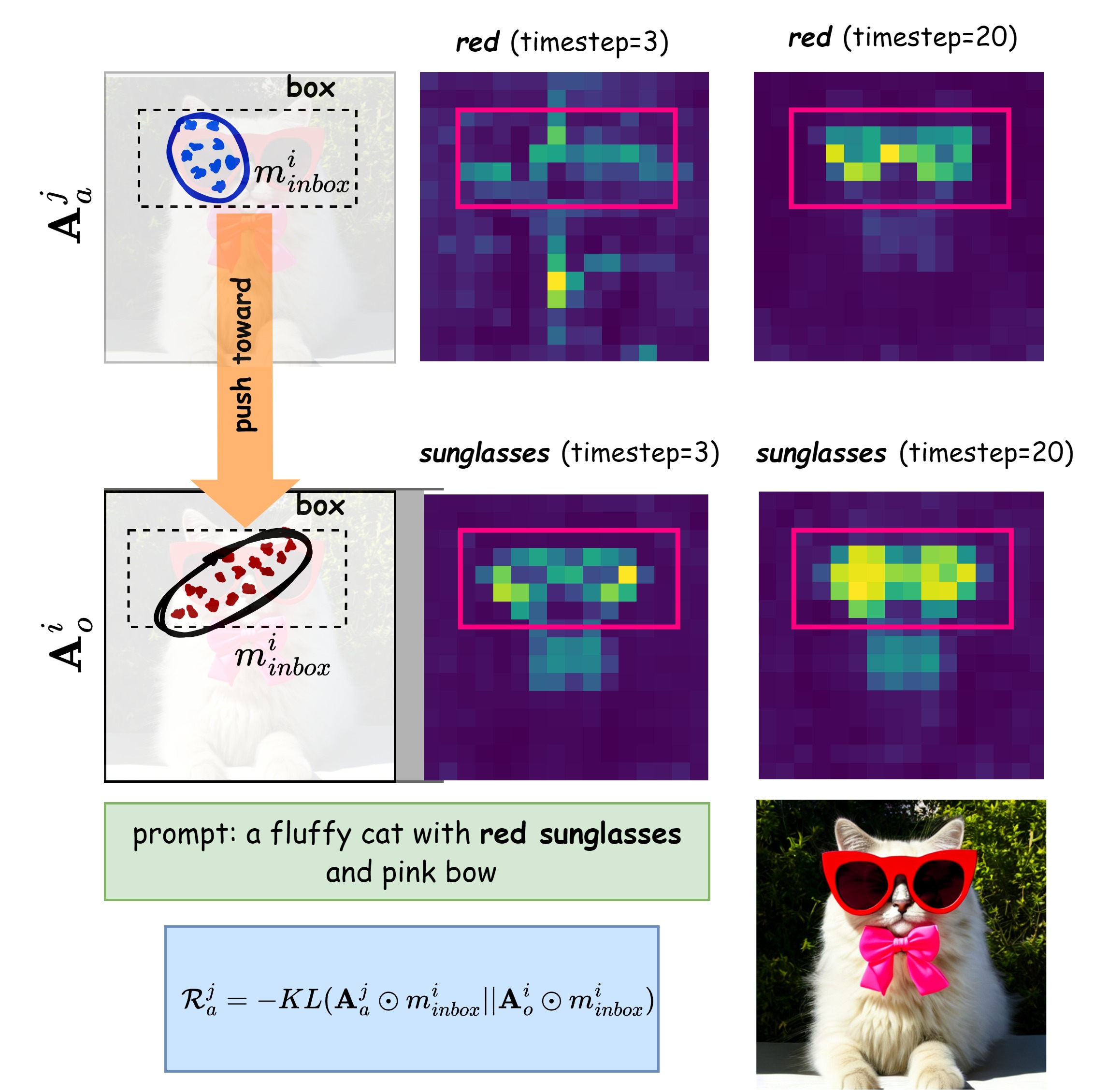}
\caption{Asymmetrical distance KL pushes attributes' distribution toward their corresponding objects' in the cross-attention maps. 
Since the attention maps of the objects are previously enriched during the \emph{generation} stage, the distribution push from attributes to their objects yields meaningful attention weights.
}
\label{fig:Bind-framework}
\end{figure}

\begin{algorithm*}
\caption{B2B Algorithm}
\label{alg:summary}
\textbf{Input:}
Input prompt $y$; LDM denoising UNet $\epsilon_{\theta}$; object and attribute token indices $\mathcal{O}$ and $\mathcal{D}$, respectively, with their corresponding mask $m_{inbox}$; total number of diffusion steps $T$; steps $\mathcal{T}_t$ where our generation and binding processes are applied.
\\\textbf{Output:} 
Refined latent $\mathbf{z}_{t}^{\prime}$ for the subsequent timestep.
%
    \begin{algorithmic}[1]
    \State  $\mathbf{A}_{S} \gets \epsilon_{\theta}(z_t, y, t)$  
    \Comment{\it{cross-attention at timestep $t$}}   
    \For{$\mathcal{T}_{t} $}
    \For{\( i\in\mathcal{O}, j\in \mathcal{D} \)}
        \State  {\it{\#object generation}}   
        \State $\mathcal{R}_{mainbox}^i \gets \mathbb{E}[\mathbf{A}_{o}^{i} \odot m_{inbox}^{i}]$
        \State $\mathcal{R}_{outbox}^i \gets \mathbb{E}[\mathbf{A}_{o}^{i} \odot (1-m_{inbox}^{i})]$
        \State $\mathcal{R}_{iou}^i \gets 1/N \sum_{k=1}^{N} $\newline
        \hspace*{6em} $IoU(\mathbf{A}_{o}^{i} \odot m_{inbox}^{i}, \mathbf{A}_{o}^{i} \odot m_{sliding}^{k} )$
        \State $\mathcal{R}_o^{i} \gets \mathcal{R}_{mainbox}^i - \mathcal{R}_{outbox}^i + \lambda_{iou} \mathcal{R}_{iou}^i$ 
        \State {\it{\# attribute binding}}    
        \State $\mathcal{R}_a^j \gets -KL(\mathbf{A}_{a}^j\odot m_{inbox}^{i}||\mathbf{A}_{o}^i\odot m_{inbox}^{i})$
        \EndFor
        \State  {\it{\# update latent \(z_t\)}}
        \State $\mathbf{z}_{t}^{\prime} \gets \mathbf{z}_{t}+\gamma \nabla\left(\sum_{i=1}^{n_o}\mathcal{R}_o^{i}+\lambda_{a}\sum_{j=1}^{n_a}\mathcal{R}_a^{j}\right)$
    \EndFor
    \end{algorithmic}
\end{algorithm*}

\subsection{Object Generation}
\label{sec::objGeneration}
To increase the probability of object generation $p\left(\mathbf{A}_{o}^i\right)$, we propose an IoU-based framework, shown in Figure~\ref{fig:IoU-framework}. From this perspective, we refer to this as the \emph{generation} stage.
Given the attention map of an object with its bounding box. The objective is to increase attention weights within the bounding box\footnote{The boolean mask $m_{inbox}$ is assigned a value of 1 for pixels inside the bounding box and 0 elsewhere.
} for the $i$-th object, i.e. $\mathbf{A}_{o}^{i} \odot m_{inbox}^{i}$, while suppressing attention weights located outside of the given bounding box, i.e. $\mathbf{A}_{o}^{i} \odot (1-m_{inbox}^{i})$. $\odot$ is an element-wise multiplication operator.
Additionally, we ensure that attention weights concentrate within the central region of the main box, thereby minimizing the dispersion of weights at its borders.
To this end, we define $N$ sliding boxes with the same dimensions as the main box. The sliding boxes are located at different positions but relatively close to the main box. 
The reward score for the object generation can be summarised as follows:
\begin{equation}
\begin{aligned}\label{eq:obj_reward}
\mathcal{R}_o^i &= \mathcal{R}_{mainbox}^i - \mathcal{R}_{outbox}^i + \lambda_{iou} \mathcal{R}_{iou}^i& 
\\& 
=\mathbb{E}[\mathbf{A}_{o}^{i} \odot m_{inbox}^{i}]
-\mathbb{E}[\mathbf{A}_{o}^{i} \odot (1-m_{inbox}^{i})]
\\&
+\frac{\lambda_{iou}}{N}\sum_{k=1}^{N} IoU(\mathbf{A}_{o}^{i} \odot m_{inbox}^{i}, \mathbf{A}_{o}^{i} \odot m_{sliding}^{k} ).
\end{aligned}
\end{equation}
$\lambda_{iou}$ is a hyperparameter for IoU rewards. 
$m_{sliding}$ denotes the sliding box situated at a distance of $(\delta_x, \delta_y)$ pixels away from the main box. 
Since the sliding boxes push objects into the center of the main box, pairs $\delta_x$ and $\delta_y$ are expected to be small. 
$N$ sliding boxes encompass different pairs $(\delta_x, \delta_y)$, where the values are randomly selected from the range of 10\% to 20\% of the minimum of the height $h$ and width $w$ of the cross-attention map.

\subsection{Attribute Binding}
\label{sec::attGeneration}
To increase the conditional probability $p\left(\mathbf{A}_{a}^j|\mathbf{A}_{o}^i\right)$, we measure how an attribute's probability distribution $p\left(\mathbf{A}_{a}^j\right)$ is different from its corresponding object's probability distribution $p(\mathbf{A}_{o}^i)$, and then decrease the distance between the given distributions. 
To this end, the Kullback–Leibler divergence (KL) is an alternative as it measures conditional dissimilarity between two probability distributions. 
We bind the $j$-th attribute to its corresponding $i$-th object via
\begin{align}\label{eq:att_reward}
     \mathcal{R}_a^j =
 -KL(\mathbf{A}_{a}^j\odot m_{inbox}^{i}||\mathbf{A}_{o}^i\odot m_{inbox}^{i}).
\end{align}
It is worthwhile to mention that the minus sign in equation \ref{eq:att_reward} ensures that equation \ref{eq:finalgrad} is satisfied for reward maximization. This maximization is obtained when the distribution of the attribute fully follows the object distribution. As shown in Figure~\ref{fig:Bind-framework}, we enforce the attributes' distribution to converge towards their respective objects. 
This is more plausible since the contents of the objects' attention maps are previously enriched in Section~\ref{sec::objGeneration}. From this perspective, we refer to it as the \emph{binding} stage.

\subsection{Implementation}
\label{sec:implementation}
Algorithm \ref{alg:summary} summarizes the pseudocode of B2B. Given a prompt $y$, along with object and attribute token indices $\mathcal{O}$ and $\mathcal{D}$ repsectively, and the respective $i^{th}$ object mask $m_{inbox}^i$, we compute the \emph{generation} and \emph{binding} rewards. These rewards update the latent encoding during specific $\mathcal{T}_{t}$ timesteps to satisfy the objective of equation \ref{eq:final}. 

%

\section{Experiments}
\label{sec::experiments}

\subsection{Experimental Setup}
\label{sec:Experiment_Setup}

\textbf{Datasets}. Several studies have utilized various prompt sets for T2I evaluation. \cite{chefer2023attendandexcite} used 66 Animal-Animal and 144 Color-Object prompts, while \cite{Xie_2023_ICCV} employed 189 prompts for single and multiple image instances. Feng \textit{et al.}~\cite{feng2023trainingfree} sampled 600 prompts from MSCOCO's 3.2K compositions \cite{lin2015microsoft}. However, to rigorously test our method, we use the T2I-CompBench \cite{huang2023t2icompbench}, which offers a comprehensive range of prompts featuring 32 colors, 23 types of textures, and 7 types of spatial relationships for evaluating attribute binding and spatial relationships. This benchmark effectively measures T2I methods' performance in attribute binding and layout generation across three categories. 

\textbf{Evaluation metrics}. We evaluate the results of our method based on the following metrics:
\begin{itemize}
\item \textbf{TIFA Score}: We employ the TIFA score \cite{hu2023tifa, sun2023dreamsync} to evaluate our method's effectiveness in generating images from textual prompts. This metric employs VQA models to determine whether questions related to the content of a generated image are accurately answered. We used a more restricted format of the TIFA score for our evaluation. 

\item \textbf{CompBench Score}: The CompBench score \cite{huang2023t2icompbench}, designed to assess fine-grained text-image correspondences, evaluates attribute binding using \textit{disentangled Blip-VQA} and spatial relationships through a \textit{UniDet-based} metric \cite{Zhou_2022_CVPR}. 
\end{itemize}
\subsection{Benchmark Results}
\label{sec:Quantit_result_sd} 
\textbf{Base Model}. We integrate the B2B approach into SD v1.4 \cite{rombach2022highresolution}, an established LDM in training-free attribute binding and layout-based studies. For each prompt, we generate 15 images. While the seeds for image generation are randomly selected, they are kept consistent across all methods to ensure a fair comparison.
\begin{table}
\centering
\begin{tabular}{@{}lcc@{}}
\toprule
T2I Models & \multicolumn{2}{c}{Color Scores} \\ \cmidrule(l){2-3} 
 & CompBench & TIFA  \\ \midrule
Stable v1-4 \cite{rombach2022highresolution} & 0.381 &  {0.312} \\ 
Stable v2 \cite{rombach2022highresolution} & 0.510 &   {0.328} \\ 
Composable \cite{liu2022compositional} & 0.417 &  {0.317}  \\ 
BoxDiff \cite{Xie_2023_ICCV} & 0.629 &  {0.339} \\ 
Structured  \cite{feng2023trainingfree} & 0.504 &  {0.326} \\ 
Att\&Exc. \cite{chefer2023attendandexcite} & 0.643 &  {0.343} \\ 
GORS \cite{huang2023t2icompbench} & 0.662 & {0.350} \\
B2B (ours) & \textbf{{0.734}} &  \textbf{0.361} \\ \bottomrule 
\end{tabular}
\caption{ Color Binding Performance Evaluated by TIFA and CompBench Scores. Best scores are shown in boldface.} 
\label{tab:color_SD} 
\end{table}
\begin{table}[t] 
\centering
\begin{tabular}{@{}lcc@{}}
\toprule
T2I Models & \multicolumn{2}{c}{Texture Scores} \\ \cmidrule(l){2-3} 
 & CompBench & TIFA  \\ \midrule
Stable v1-4 \cite{rombach2022highresolution} & 0.418 & {0.318} \\
Stable v2 \cite{rombach2022highresolution} & 0.498 & {0.324}  \\
Composable \cite{liu2022compositional} & 0.375 & {0.306} \\
BoxDiff \cite{Xie_2023_ICCV} & 0.591 & {0.325} \\
Structured  \cite{feng2023trainingfree} & 0.494 & {0.321} \\
Att\&Exc. \cite{chefer2023attendandexcite} & 0.605 & {0.328} \\
GORS \cite{huang2023t2icompbench} & 0.630 & {0.337} \\
B2B (ours) & \textbf{{0.689}} & \textbf{0.353} \\ \bottomrule
\end{tabular}
\caption{Texture Binding Performance Evaluated by TIFA and CompBench Scores.}
\label{tab:texture_SD}
\end{table}
%
%

\textbf{Baselines.} We compare B2B with various conventional and state-of-the-art attribute binding and layout-based techniques. For attribute binding, we benchmark ourselves against both training-free methods, including Attend-and-Excite \cite{chefer2023attendandexcite} and Structured Diffusion \cite{feng2023trainingfree}, as well as fine-tuning-based methods like GORS \cite{huang2023t2icompbench}. 
In terms of layout-based methods, we evaluate our approach alongside the recently introduced BoxDiff \cite{Xie_2023_ICCV} method.

\textbf{Quantitative results.} Table \ref{tab:color_SD} reports the color binding results of the techniques in terms of  CompBench and TIFA scores where B2B achieves the highest score in color binding by a considerable margin compared to the others. 
B2B notably outperforms the training-free Attend-and-Excite and the fine-tuning-based GORS method. The strict TIFA score, requiring all VQA-answered questions from the generated image to be accurate, leads to a more modest improvement in B2B's color accuracy. This modesty is due to the strict version of the TIFA score we used, in contrast to the broader improvements reflected in the CompBench score. 
Table \ref{tab:texture_SD} presents texture binding results of the methods, highlighting the increased difficulty compared to the color binding task, as indicated by generally lower scores. 
Similar to color binding results, B2B outperforms methods such as Attend-and-Excite and GORS by a high margin.
In Table \ref{tab:spatial_SD}, we reported results for spatial reasoning assessment. CompBench, as an explainable metric, evaluates the generated contents based on the comparison of detected bounding boxes for each object, while TIFA relies on question-answer pairs from a VQA model. Our method exhibits clear superiority in both metrics. The results also suggest significant room for improvement in spatial reasoning.

\begin{figure*}[t] 
\centering
\includegraphics[width=\textwidth]{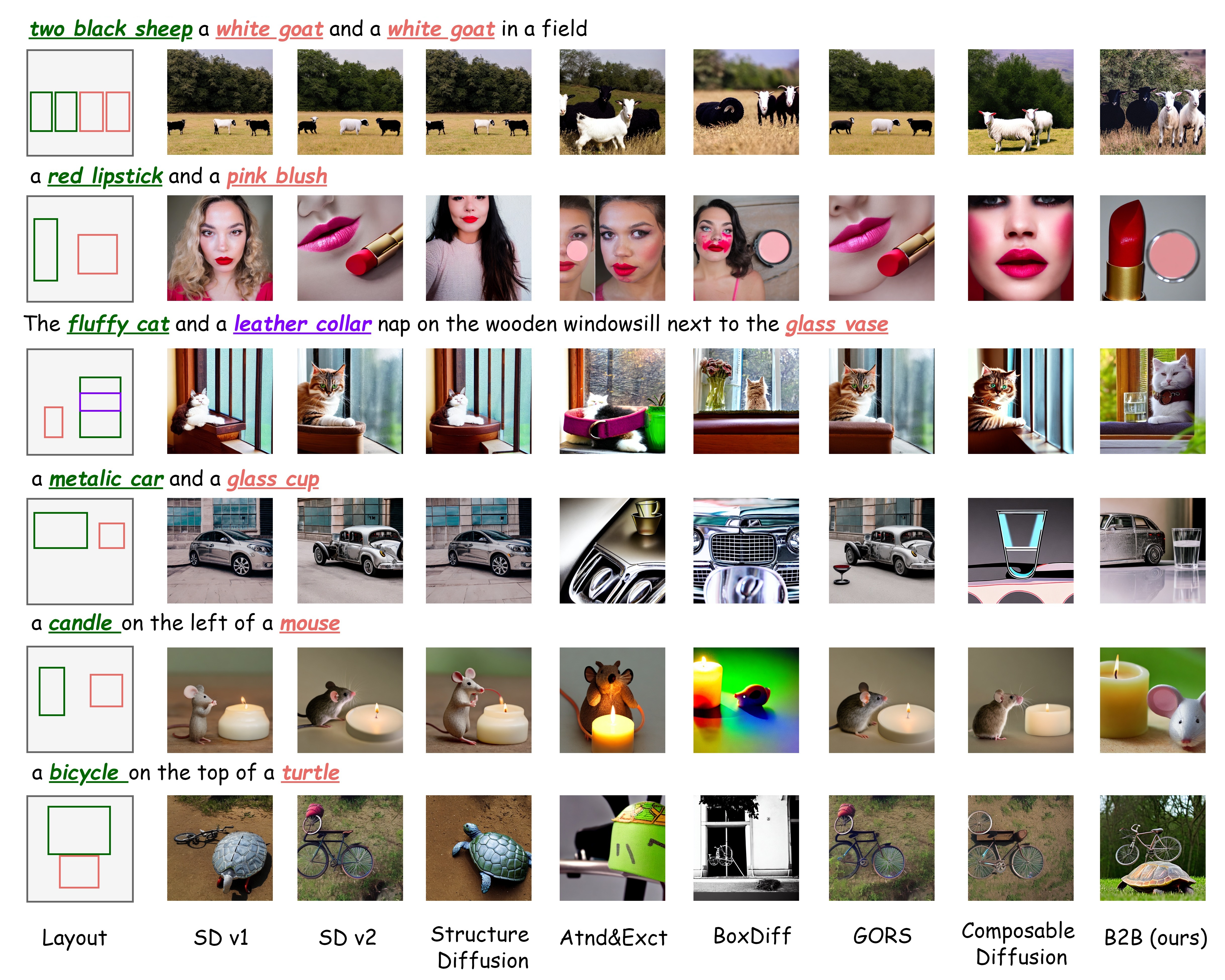} 
\caption{Visual comparison of methods for different scenarios, including color binding, texture binding, and spatial reasoning.}
\label{fig:image-b2b-competitors}
\end{figure*}

\textbf{Qualitative Results}. Figure \ref{fig:image-b2b-competitors} displays samples generated by different approaches. B2B consistently adheres to the attributes specified in the prompts while maintaining the intended layout. 
Particularly noteworthy is the second row, where B2B successfully binds object tokens like `lipstick' and `blush' to their designated locations and corresponding colors, while the other methods often produce biased results, such as facial images typical in cosmetic applications. 
With texture-related prompts, B2B demonstrates fidelity to both the specified layout and textures. This is more evident in the third row, where the generated image adheres to the layout, avoids catastrophic neglect, and correctly binds to the texture attributes, even when faced with a lengthy prompt. 
The fifth and sixth rows showcase the spatial reasoning capabilities of our method. B2B accurately follows the specified layout, which is crucial for spatial reasoning, as evident in the unusual compositions of the sixth row. 
\begin{table}[t]
\centering
\begin{tabular}{@{}lcc@{}}
\toprule
T2I Models & \multicolumn{2}{c}{Spatial Scores} \\ \cmidrule(l){2-3} 
 & CompBench & TIFA  \\ \midrule
Stable v1-4 \cite{rombach2022highresolution} & 0.125 & {0.108} \\ 
Stable v2 \cite{rombach2022highresolution} & 0.136 & {0.113} \\ 
Composable \cite{liu2022compositional} & 0.108 & {0.091} \\ 
BoxDiff \cite{Xie_2023_ICCV} & {0.228} & {0.171} \\ 
Structured  \cite{feng2023trainingfree} & 0.147 & {0.122} \\ 
Att\&Exc. \cite{chefer2023attendandexcite} & 0.151 &  {0.127} \\ 
GORS \cite{huang2023t2icompbench} & 0.182 & {0.145} \\ 
B2B (ours) & \textbf{0.292} & \textbf{0.214} \\ \bottomrule 
\end{tabular}
\caption{Spatial Reasoning Performance Evaluated by TIFA and CompBench Scores.} 
\label{tab:spatial_SD} 
\end{table}

\subsection{Plug-and-Play Analysis}\label{sec:plug_analys}

\textbf{Base Model}. Our method's plug-and-play capability is evaluated using the GLIGEN model \cite{li2023gligen} that includes a gated attention (GA) module for conditioning on inputs like bounding boxes. GA facilitates the generation of objects within specific bounding box coordinates, and we enhance its performance with our \emph{generation} and \emph{binding} modules in its $16 \times 16$ resolution cross-attention layers.

\begin{figure*}
\centering
\includegraphics[width=.8\columnwidth]{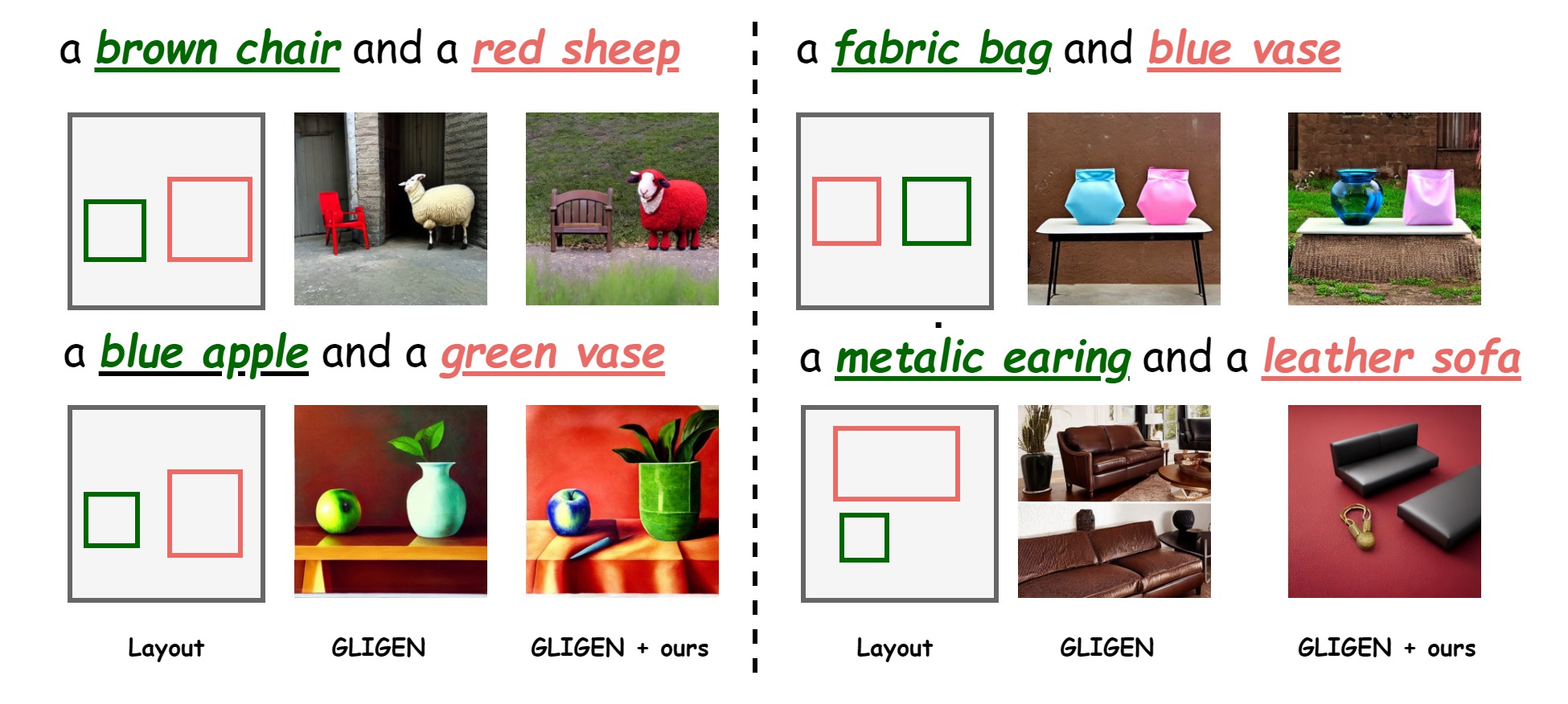} 
\caption{Qualitative results demonstrating the plug-and-play effectiveness of our method on GLIGEN's color (left column) and texture (right column) binding. Zoom in for a clearer view.}
\label{fig:b2b_gligen}
\end{figure*}

\textbf{Binding Analysis}.
We assessed GLIGEN's attribute binding using Compbench and TIFA scores. The integration of B2B with GLIGEN significantly improved color and texture binding, as shown in Table \ref{tab:glig_attrib}, demonstrating our method's effectiveness and plug-and-play adaptability. 
The qualitative results are presented in Figure \ref{fig:b2b_gligen}. Integrating the B2B module into GLIGEN enhances color and texture binding, simultaneously contributing to the mitigation of catastrophic neglect. 
This improvement is particularly noticeable in the case of the `metallic earring' (second row) in the image generated using GLIGEN, where neglect is effectively addressed.

\begin{table}
\centering
\resizebox{0.6\columnwidth}{!}{%
\begin{tabular}{@{}lcccc@{}}
\toprule
Methods & \multicolumn{2}{c}{CompBench score} & \multicolumn{2}{c}{TIFA score} \\
\cmidrule(lr){2-3} \cmidrule(lr){4-5}
 & Color & Texture & Color & Texture \\
\midrule
GLIGEN~\cite{li2023gligen} & 0.406 & 0.425 & 0.315 & 0.319 \\
GLIGEN + B2B (Ours) & \textbf{0.692} & \textbf{{0.637}} & \textbf{0.354} & \textbf{0.335} \\
\bottomrule
\end{tabular}%
}
\caption{Plug-and-play effectiveness of our proposed method on GLIGEN attribute binding.}
\label{tab:glig_attrib}
\end{table}
%

\begin{figure}
\centering
\includegraphics[width=0.65\columnwidth]{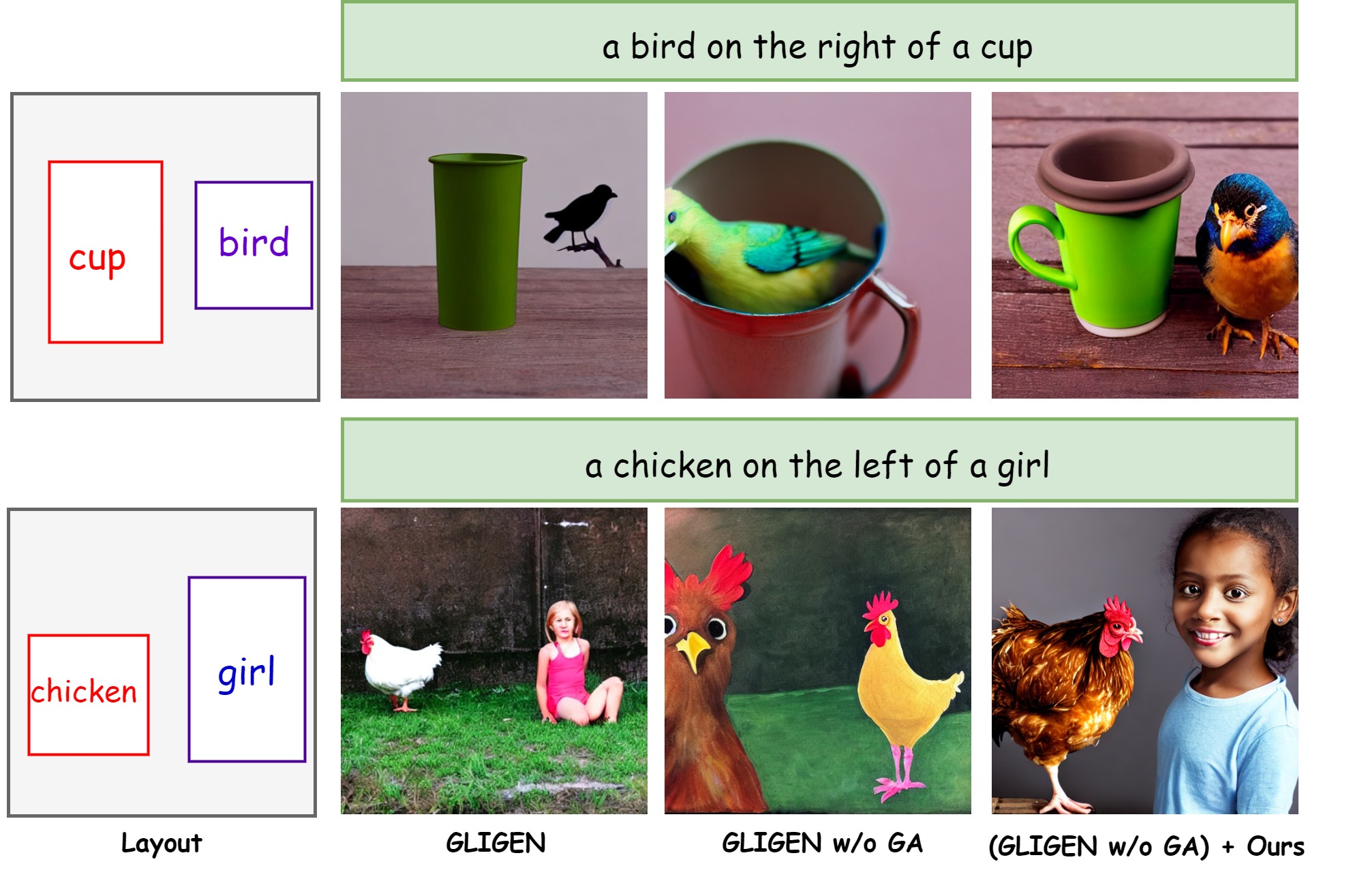}
\caption{Qualitative analysis of removing and replacing the GA in the GLIGEN model with \emph{generation} module: effects on visual quality and spatial reasoning}
\label{fig:fid_analysis_main}
\end{figure}

\textbf{Spatial Analysis}.
GLIGEN is specialized by its GA module and trained on bounding box-annotated data that excels in layout adherence, hence enhancing spatial reasoning. However, this leads to a compromise in image quality \cite{li2023gligen,lee2024parrot}. 
We explored GLIGEN's architecture in three scenarios to optimize spatial reasoning and visual quality:
a) Original GLIGEN network's spatial and FID\footnote{FID metric \cite{heusel2018gans} is used for visual quality assessment.} scores evaluation. 
b) Spatial and FID scores assessment after removing GLIGEN's GA module.
c) Replacing GLIGEN's GA module with our \emph{generation} module to analyze spatial and FID scores. {Table} \ref{tab:gated_atten_analys} reveals that removing the GA reduces the FID score, indicating an improvement in visual quality, but at the expense of spatial reasoning capabilities (refer to {Figure} \ref{fig:fid_analysis_main}). Conversely, substituting the GA with our training-free \emph{generation} module leads to a low FID score, thereby enhancing visual quality, and maintains spatial reasoning (as shown in Figure \ref{fig:fid_analysis_main}). This is significant given that the GA module, unlike our module, requires training on additional annotated data. 

\begin{figure*}
\centering
\includegraphics[width=0.8\columnwidth]{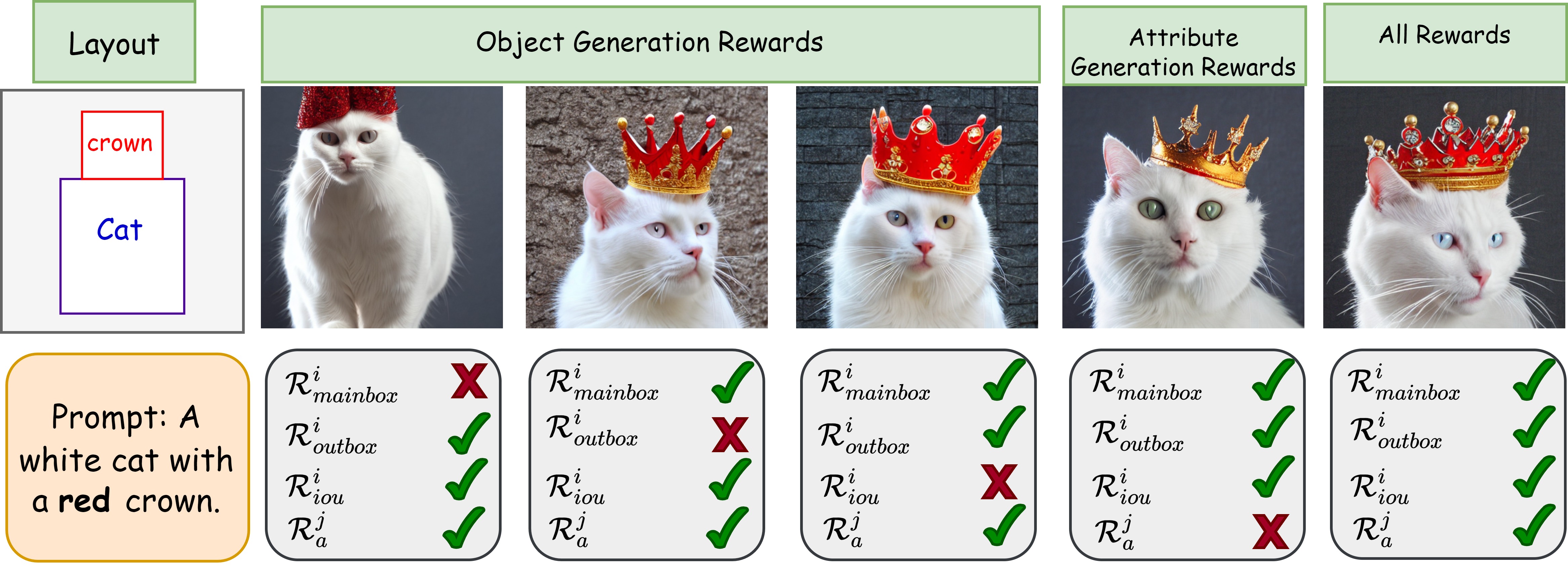}
\caption{Visual ablation study on the reward elements of B2B.}
\label{fig:vis_abl_main}
\end{figure*}

\begin{table*}
\centering
\resizebox{0.75\columnwidth}{!}{%
\begin{tabular}{@{}lccc@{}}
\toprule
GLIGEN & \multicolumn{1}{c}{CompBench score} & \multicolumn{1}{c}{TIFA score} & \multicolumn{1}{c}{FID score} \\
\cmidrule(lr){2-2} \cmidrule(lr){3-3} 
 & Spatial & Spatial & \\
\midrule
w Gated-Attention  & \textbf{0.361} & \textbf{0.256} & \textbf{20.62} \\
w/o Gated-Attention & {0.128} & {0.110} & 20.29 \\ 
w/o Gated-Attention + B2B (Ours) & {{0.298}} & {0.215} & {20.32} \\
\bottomrule
\end{tabular}%
}
\caption{Quantitative analysis of the plug-and-play effectiveness of our \emph{generation} module integrated into the GLIGEN model.}
\label{tab:gated_atten_analys}
\end{table*}


\subsection{Ablation Study}
\label{sec:ablation} 
In our ablation study, we selected 50 random prompts from each category of color, texture, and spatial benchmarks, generating five images for each prompt. The outcomes of this quantitative analysis, assessed by CompBench, are tabulated in {Tables} \ref{tab:main_spat_abs} and \ref{tab:main_attrib_abl}. 
Figure~\ref{fig:vis_abl_main} also depicts a visual ablation study. 
The results indicate the critical role of generation rewards in spatial reasoning and adhering to the layout for object placement. The $\mathcal{R}_{mainbox}^{i}$, as illustrated with the 'crown' example in Figure \ref{fig:vis_abl_main}, effectively counters catastrophic neglect. 
Meanwhile, the $\mathcal{R}_{outbox}^{i}$ facilitates approximate alignment of objects with the intended layout, and the $\mathcal{R}_{iou}^{i}$ ensures exact positioning within specific bounding boxes, thus improving both the visual appeal and adherence to the layout. Additionally, the importance of the $\mathcal{R}_{a}^{j}$ in attribute binding is highlighted, as its removal results in improper color binding of objects.

\begin{table}
\centering
\small 
\begin{tabular}{@{}cccc@{}}
\toprule
\multicolumn{3}{c}{Reward Components} & \multicolumn{1}{c}{CompBench score} \\
\cmidrule(lr){1-3} \cmidrule(lr){4-4}
\(\mathcal{R}_{mainbox}^{i}\) & \(\mathcal{R}_{outbox}^{i}\) & \(\mathcal{R}_{iou}^{i}\) & Spatial \\
\midrule
\(\times\) & \(\times\) & \(\times\) & 0.124 \\
\checkmark & \(\times\) & \(\times\) & 0.167 \\
\(\times\) & \checkmark & \(\times\) & 0.143 \\
\(\times\) & \(\times\) & \checkmark & 0.131 \\
\checkmark & \checkmark & \checkmark & 0.289 \\
\bottomrule
\end{tabular}
\caption{Ablation on various elements of the object generation reward (equation~\ref{eq:obj_reward}).}
\label{tab:main_spat_abs}
\end{table}

\begin{table}
\centering
\small 
\begin{tabular}{@{}cccccc@{}}
\toprule
\multicolumn{4}{c}{Reward Components} & \multicolumn{2}{c}{CompBench score} \\
\cmidrule(lr){1-4} \cmidrule(lr){5-6}
\(\mathcal{R}_{mainbox}^{i}\) & \(\mathcal{R}_{outbox}^{i}\) & \(\mathcal{R}_{iou}^{i}\) &  \(\mathcal{R}_{a}^{j}\) & Color & Texture \\
\midrule
\(\times\) & \(\times\) & \(\times\) & \(\times\) & 0.379 & 0.416 \\
\(\times\) & \checkmark & \checkmark & \(\times\) & 0.391 & 0.429 \\
\checkmark & \checkmark & \checkmark & \(\times\) & 0.622 & 0.589 \\
\(\times\) & \checkmark & \checkmark & \checkmark & 0.582 & 0.529 \\
\checkmark & \checkmark & \checkmark & \checkmark & 0.731 & 0.681 \\
\bottomrule
\end{tabular}
\caption{Ablation study on various combinations of object generation and attribute generation reward components.}
\label{tab:main_attrib_abl}
\end{table}

\section{Conclusion}
\label{sec:conclusion}
In this study, we introduced the B2B model to tackle major challenges in text-to-image (T2I) Latent Diffusion Models (LDMs), focusing on attribute binding and spatial control. B2B employs a dual-module system, \emph{Generation} and \emph{Binding}, to effectively address catastrophic neglect, improve attribute binding precision, and ensure accurate object placement. Its compatibility as a plug-and-play module with existing T2I frameworks is demonstrated through its outstanding performance in CompBench and TIFA benchmarks, signifying a major leap in generative modeling. B2B's breakthroughs highlight its role as a potential standard for future research, paving the way for innovative developments in digital imaging and generative AI.

\bibliographystyle{unsrt}  
\bibliography{references}

\begin{thebibliography}{10}

\bibitem{rombach2022highresolution}
Robin Rombach, Andreas Blattmann, Dominik Lorenz, Patrick Esser, and Björn Ommer.
\newblock High-resolution image synthesis with latent diffusion models, 2022.

\bibitem{li2023gligen}
Yuheng Li, Haotian Liu, Qingyang Wu, Fangzhou Mu, Jianwei Yang, Jianfeng Gao, Chunyuan Li, and Yong~Jae Lee.
\newblock Gligen: Open-set grounded text-to-image generation.
\newblock {\em CVPR}, 2023.

\bibitem{mardani2023variational}
Morteza Mardani, Jiaming Song, Jan Kautz, and Arash Vahdat.
\newblock A variational perspective on solving inverse problems with diffusion models, 2023.

\bibitem{sun2023dreamsync}
Jiao Sun, Deqing Fu, Yushi Hu, Su~Wang, Royi Rassin, Da-Cheng Juan, Dana Alon, Charles Herrmann, Sjoerd van Steenkiste, Ranjay Krishna, and Cyrus Rashtchian.
\newblock Dreamsync: Aligning text-to-image generation with image understanding feedback, 2023.

\bibitem{peng2023metadataconditioned}
Wei Peng, Tomas Bosschieter, Jiahong Ouyang, Robert Paul, Ehsan Adeli, Qingyu Zhao, and Kilian~M. Pohl.
\newblock Metadata-conditioned generative models to synthesize anatomically-plausible 3d brain mris, 2023.

\bibitem{Feng_2023_ICCV}
Chun-Mei Feng, Kai Yu, Yong Liu, Salman Khan, and Wangmeng Zuo.
\newblock Diverse data augmentation with diffusions for effective test-time prompt tuning.
\newblock In {\em Proceedings of the IEEE/CVF International Conference on Computer Vision (ICCV)}, pages 2704--2714, October 2023.

\bibitem{huang2024creativesynth}
Nisha Huang, Weiming Dong, Yuxin Zhang, Fan Tang, Ronghui Li, Chongyang Ma, Xiu Li, and Changsheng Xu.
\newblock Creativesynth: Creative blending and synthesis of visual arts based on multimodal diffusion, 2024.

\bibitem{yang2024mastering}
Ling Yang, Zhaochen Yu, Chenlin Meng, Minkai Xu, Stefano Ermon, and Bin Cui.
\newblock Mastering text-to-image diffusion: Recaptioning, planning, and generating with multimodal llms, 2024.

\bibitem{yu2022scaling}
Jiahui Yu, Yuanzhong Xu, Jing~Yu Koh, Thang Luong, Gunjan Baid, Zirui Wang, Vijay Vasudevan, Alexander Ku, Yinfei Yang, Burcu~Karagol Ayan, Ben Hutchinson, Wei Han, Zarana Parekh, Xin Li, Han Zhang, Jason Baldridge, and Yonghui Wu.
\newblock Scaling autoregressive models for content-rich text-to-image generation, 2022.

\bibitem{chen2023trainingfree}
Minghao Chen, Iro Laina, and Andrea Vedaldi.
\newblock Training-free layout control with cross-attention guidance, 2023.

\bibitem{kim2023dense}
Yunji Kim, Jiyoung Lee, Jin-Hwa Kim, Jung-Woo Ha, and Jun-Yan Zhu.
\newblock Dense text-to-image generation with attention modulation, 2023.

\bibitem{lee2024compose}
Jonghyun Lee, Hansam Cho, Youngjoon Yoo, Seoung~Bum Kim, and Yonghyun Jeong.
\newblock Compose and conquer: Diffusion-based 3d depth aware composable image synthesis, 2024.

\bibitem{wang2024instantid}
Qixun Wang, Xu~Bai, Haofan Wang, Zekui Qin, and Anthony Chen.
\newblock Instantid: Zero-shot identity-preserving generation in seconds.
\newblock {\em arXiv preprint arXiv:2401.07519}, 2024.

\bibitem{lee2024parrot}
Seung~Hyun Lee, Yinxiao Li, Junjie Ke, Innfarn Yoo, Han Zhang, Jiahui Yu, Qifei Wang, Fei Deng, Glenn Entis, Junfeng He, Gang Li, Sangpil Kim, Irfan Essa, and Feng Yang.
\newblock Parrot: Pareto-optimal multi-reward reinforcement learning framework for text-to-image generation, 2024.

\bibitem{phung2023grounded}
Quynh Phung, Songwei Ge, and Jia-Bin Huang.
\newblock Grounded text-to-image synthesis with attention refocusing, 2023.

\bibitem{zhang2023adding}
Lvmin Zhang, Anyi Rao, and Maneesh Agrawala.
\newblock Adding conditional control to text-to-image diffusion models, 2023.

\bibitem{feng2023trainingfree}
Weixi Feng, Xuehai He, Tsu-Jui Fu, Varun Jampani, Arjun~Reddy Akula, Pradyumna Narayana, Sugato Basu, Xin~Eric Wang, and William~Yang Wang.
\newblock Training-free structured diffusion guidance for compositional text-to-image synthesis.
\newblock In {\em The Eleventh International Conference on Learning Representations}, 2023.

\bibitem{mo2023freecontrol}
Sicheng Mo, Fangzhou Mu, Kuan~Heng Lin, Yanli Liu, Bochen Guan, Yin Li, and Bolei Zhou.
\newblock Freecontrol: Training-free spatial control of any text-to-image diffusion model with any condition, 2023.

\bibitem{zhao2023loco}
Peiang Zhao, Han Li, Ruiyang Jin, and S.~Kevin Zhou.
\newblock Loco: Locally constrained training-free layout-to-image synthesis, 2023.

\bibitem{yu2023freedom}
Jiwen Yu, Yinhuai Wang, Chen Zhao, Bernard Ghanem, and Jian Zhang.
\newblock Freedom: Training-free energy-guided conditional diffusion model.
\newblock {\em Proceedings of the IEEE/CVF International Conference on Computer Vision (ICCV)}, 2023.

\bibitem{chefer2023attendandexcite}
Hila Chefer, Yuval Alaluf, Yael Vinker, Lior Wolf, and Daniel Cohen-Or.
\newblock Attend-and-excite: Attention-based semantic guidance for text-to-image diffusion models, 2023.

\bibitem{huang2023t2icompbench}
Kaiyi Huang, Kaiyue Sun, Enze Xie, Zhenguo Li, and Xihui Liu.
\newblock T2i-compbench: A comprehensive benchmark for open-world compositional text-to-image generation, 2023.

\bibitem{hu2023tifa}
Yushi Hu, Benlin Liu, Jungo Kasai, Yizhong Wang, Mari Ostendorf, Ranjay Krishna, and Noah~A Smith.
\newblock Tifa: Accurate and interpretable text-to-image faithfulness evaluation with question answering, 2023.

\bibitem{gupta2023photorealistic}
Agrim Gupta, Lijun Yu, Kihyuk Sohn, Xiuye Gu, Meera Hahn, Li~Fei-Fei, Irfan Essa, Lu~Jiang, and José Lezama.
\newblock Photorealistic video generation with diffusion models, 2023.

\bibitem{basu2023editval}
Samyadeep Basu, Mehrdad Saberi, Shweta Bhardwaj, Atoosa~Malemir Chegini, Daniela Massiceti, Maziar Sanjabi, Shell~Xu Hu, and Soheil Feizi.
\newblock Editval: Benchmarking diffusion based text-guided image editing methods, 2023.

\bibitem{Zeng_2023_CVPR}
Yu~Zeng, Zhe Lin, Jianming Zhang, Qing Liu, John Collomosse, Jason Kuen, and Vishal~M. Patel.
\newblock Scenecomposer: Any-level semantic image synthesis.
\newblock In {\em Proceedings of the IEEE/CVF Conference on Computer Vision and Pattern Recognition (CVPR)}, pages 22468--22478, June 2023.

\bibitem{gani2023llm}
Hanan Gani, Shariq~Farooq Bhat, Muzammal Naseer, Salman Khan, and Peter Wonka.
\newblock Llm blueprint: Enabling text-to-image generation with complex and detailed prompts, 2023.

\bibitem{wu2023selfcorrecting}
Tsung-Han Wu, Long Lian, Joseph~E. Gonzalez, Boyi Li, and Trevor Darrell.
\newblock Self-correcting llm-controlled diffusion models, 2023.

\bibitem{zhang2023controllable}
Tianjun Zhang, Yi~Zhang, Vibhav Vineet, Neel Joshi, and Xin Wang.
\newblock Controllable text-to-image generation with gpt-4, 2023.

\bibitem{lian2023llmgrounded}
Long Lian, Boyi Li, Adam Yala, and Trevor Darrell.
\newblock Llm-grounded diffusion: Enhancing prompt understanding of text-to-image diffusion models with large language models.
\newblock {\em arXiv preprint arXiv:2305.13655}, 2023.

\bibitem{liu2022compositional}
Nan Liu, Shuang Li, Yilun Du, Antonio Torralba, and Joshua~B Tenenbaum.
\newblock Compositional visual generation with composable diffusion models.
\newblock In {\em Computer Vision--ECCV 2022: 17th European Conference, Tel Aviv, Israel, October 23--27, 2022, Proceedings, Part XVII}, pages 423--439. Springer, 2022.

\bibitem{li2023divide}
Yumeng Li, Margret Keuper, Dan Zhang, and Anna Khoreva.
\newblock Divide \& bind your attention for improved generative semantic nursing, 2023.

\bibitem{rassin2023linguistic}
Royi Rassin, Eran Hirsch, Daniel Glickman, Shauli Ravfogel, Yoav Goldberg, and Gal Chechik.
\newblock Linguistic binding in diffusion models: Enhancing attribute correspondence through attention map alignment.
\newblock In {\em Thirty-seventh Conference on Neural Information Processing Systems}, 2023.

\bibitem{bartal2023multidiffusion}
Omer Bar-Tal, Lior Yariv, Yaron Lipman, and Tali Dekel.
\newblock Multidiffusion: Fusing diffusion paths for controlled image generation, 2023.

\bibitem{liu2023syncdreamer}
Yuan Liu, Cheng Lin, Zijiao Zeng, Xiaoxiao Long, Lingjie Liu, Taku Komura, and Wenping Wang.
\newblock Syncdreamer: Generating multiview-consistent images from a single-view image, 2023.

\bibitem{Xie_2023_ICCV}
Jinheng Xie, Yuexiang Li, Yawen Huang, Haozhe Liu, Wentian Zhang, Yefeng Zheng, and Mike~Zheng Shou.
\newblock Boxdiff: Text-to-image synthesis with training-free box-constrained diffusion.
\newblock In {\em Proceedings of the IEEE/CVF International Conference on Computer Vision (ICCV)}, pages 7452--7461, 2023.

\bibitem{rombach2022high}
Robin Rombach, Andreas Blattmann, Dominik Lorenz, Patrick Esser, and Bj{\"o}rn Ommer.
\newblock High-resolution image synthesis with latent diffusion models.
\newblock In {\em Proceedings of the IEEE/CVF conference on computer vision and pattern recognition}, pages 10684--10695, 2022.

\bibitem{ho2020denoising}
Jonathan Ho, Ajay Jain, and Pieter Abbeel.
\newblock Denoising diffusion probabilistic models.
\newblock {\em Advances in neural information processing systems}, 33:6840--6851, 2020.

\bibitem{yuan2023rewarddirected}
Hui Yuan, Kaixuan Huang, Chengzhuo Ni, Minshuo Chen, and Mengdi Wang.
\newblock Reward-directed conditional diffusion: Provable distribution estimation and reward improvement.
\newblock In {\em Thirty-seventh Conference on Neural Information Processing Systems}, 2023.

\bibitem{huang2023diffusion}
Tao Huang, Guangqi Jiang, Yanjie Ze, and Huazhe Xu.
\newblock Diffusion reward: Learning rewards via conditional video diffusion, 2023.

\bibitem{lin2015microsoft}
Tsung-Yi Lin, Michael Maire, Serge Belongie, Lubomir Bourdev, Ross Girshick, James Hays, Pietro Perona, Deva Ramanan, C.~Lawrence Zitnick, and Piotr Dollár.
\newblock Microsoft coco: Common objects in context, 2015.

\bibitem{Zhou_2022_CVPR}
Xingyi Zhou, Vladlen Koltun, and Philipp Kr\"ahenb\"uhl.
\newblock Simple multi-dataset detection.
\newblock In {\em Proceedings of the IEEE/CVF Conference on Computer Vision and Pattern Recognition (CVPR)}, pages 7571--7580, June 2022.

\bibitem{heusel2018gans}
Martin Heusel, Hubert Ramsauer, Thomas Unterthiner, Bernhard Nessler, and Sepp Hochreiter.
\newblock Gans trained by a two time-scale update rule converge to a local nash equilibrium, 2018.

\end{thebibliography}

\end{document}